\documentclass[10pt,twocolumn,letterpaper]{article}

\usepackage{iccv}
\usepackage{times}
\usepackage{epsfig}
\usepackage{graphicx}
\usepackage{amsmath}
\usepackage{amssymb}
\usepackage{bbm}
\usepackage{bm}
\usepackage{booktabs}
\usepackage{multirow}
\usepackage{threeparttable}
\usepackage{algorithm}
\usepackage{listings}
\usepackage{algpseudocode}  
\usepackage{amsmath}  
\usepackage{enumitem}
\usepackage{comment}

\usepackage[breaklinks=true,bookmarks=false]{hyperref}

\iccvfinalcopy 

\def\iccvPaperID{****} 

\ificcvfinal\fi

\begin{document}

\title{Large-Scale Unsupervised Person Re-Identification with Contrastive Learning}
\author{Weiquan Huang$^*$, Yan Bai\thanks{Equal Contribution.} , Qiuyu Ren, Xinbo Zhao, Ming Feng and Yin Wang\\
Tongji University\\
Shanghai, China\\
{\tt\small \{weiquanh,yan.bai,qiuyu.ren,1930792,1810865,yinw\}@tongji.edu.cn}
}






\maketitle
\ificcvfinal\thispagestyle{empty}\fi

\begin{abstract}
Existing public person Re-Identification~(ReID) datasets are small in modern terms because of labeling difficulty.
Although unlabeled surveillance video is abundant and relatively easy to obtain, it is unclear how to leverage these footage to learn meaningful ReID representations.
In particular, most existing unsupervised and domain adaptation ReID methods utilize only the public datasets in their experiments, with labels removed.
In addition, due to small data sizes, these methods usually rely on fine tuning by the unlabeled training data in the testing domain to achieve good performance.

Inspired by the recent progress of large-scale self-supervised image classification using contrastive learning, we propose to learn ReID representation from large-scale unlabeled surveillance video alone.
Assisted by off-the-shelf pedestrian detection tools, we apply the contrastive loss at both the image and the tracklet levels.
Together with a principal component analysis step using camera labels freely available, our evaluation using a  large-scale unlabeled dataset shows far superior performance among unsupervised methods that do not use any training data in the testing domain.
Furthermore, the accuracy improves with the data size and therefore our method has great potential with even larger and more diversified datasets.
\end{abstract}

\section{Introduction}\label{sec:intro}

\begin{table}[ht!]
\centering
\small
\setlength\tabcolsep{2pt} 
\begin{threeparttable}
\begin{tabular}{c|cccccc}
\hline
Dataset       & label & ID    & Cam  & BBox  & tracklet & hour\\ \hline
Market-1501   & yes & 751 & 6   & 12,936 & - & - \\ 
DukeMTMC-reID & yes & 702 & 8 & 15,820  & - & - \\ 
MSMT17-V2     & yes & 1,041 & 15 & 30,248 & - & - \\  
\midrule
MARS          & yes & 625 & 6 & 509,914 & 8,298 & - \\  
DukeMTMC-Video & yes & 702 & 8 & 369,656 & 2,196 & 85~m \\ 
CampusT~(ours) & no & 11K\tnote{*} & 24 & 15.4M\tnote{\textdagger} & 946K\tnote{\textdaggerdbl} & 480~h \\ \hline
\end{tabular}
\begin{tablenotes}
\footnotesize
\item[*] based on the student population of the closed campus. Since the dataset spans over five days, the effective IDs could be more if we consider a person with different clothes as different IDs.
\item[\textdagger] off-the-shelf pedestrian detection tool applied at 2~Hz.
\item[\textdaggerdbl] including only tracklets longer than two seconds based on the algorithm described in Section~\ref{sec:method-tracklet}.
\end{tablenotes}
\end{threeparttable}
\vspace{1ex}
\caption{Comparison with the largest public person ReID datasets (training data), which are very homogeneous. The first section is image-based datasets and the second is video based.}
\label{tab:datasets}
\end{table}

The research on person re-identification~(ReID) has greatly benefited from a number of datasets widely used as benchmarks in the literature~\cite{zheng2015scalable_market,zheng2017unlabeled_dukereid,ristani2016MTMC,Zheng2016MARSAV}.
After five years, however, these datasets are small by modern standards and have arguably limited further advances.
For example, existing supervised ReID methods are approaching the accuracy limit even for the challenging Rank-1 benchmark~\cite{Luo2019bag,zheng2019joint_dgnet,zhu2020viewpoint}.
Unsupervised ReID and unsupervised domain adaptation ReID methods are quickly narrowing the gap toward full supervision performance by clever clustering and pseudo label assignment algorithms that are very effective against these datasets~\cite{ge2020mutual,wu2019unsupervised_UGA,zhu2019intra_MTML,zeng2020hierarchical_HCT}.

With the inherent difficulty of ReID labeling, it is unclear whether we can significantly scale the labeled data.
Table~\ref{tab:datasets} shows the statistics of the three largest image ReID datasets and the two largest video ReID datasets.
The total ID, camera, and image/tracklet counts are very similar, which may indicate the memory limitation of human labeling.

On the other hand, surveillance cameras are prevalent nowadays and the raw footage is relatively easy to obtain.
The last row of Table~\ref{tab:datasets} is 480 hours of footage from 24 cameras in a university campus which is a small portion of her surveillance network.
Although this dataset dwarfs the existing ones, it is unclear how to leverage unlabeled ReID data of this scale.
Existing unsupervised ReID methods in the literature perform experiments on the aforementioned public datasets almost exclusively.
Furthermore, these methods usually utilize the training data of the testing domain (with labels removed) to achieve good performance.
The direct transfer of a model trained by a different domain performs poorly~\cite{fu2019self,wang2020cycas}.

Inspired by the recent advances of self-supervised image classification using contrastive learning~\cite{he2020momentum_moco,chen2020big_simclrv2}, we propose to learn ReID representation from large-scale unlabeled pedestrian video only.
First we use Detectron2~\cite{wu2019detectron2} to extract person bounding boxes from video frames and apply contrastive learning on these images to learn a basic ReID representation.
Observing that contrastive learning against individual images performs poorly, consistent with another report~\cite{ge2020self}, we enlarge the positive datasets with images of consecutive frames using the \emph{mutual nearest neighbor}~(MNN) criteria, which is similar to \emph{tracklets} frequently used in unsupervised ReID methods~\cite{li2020unsupervised_UTAL}.

A tracklet is a sequence of images of a person traveling through a camera view.
Our tracklet usage is much weaker than those in the literature because we do not associate tracklets of the same ID within or across cameras.
Even one passage of a person under a camera is usually broken into multiple segments due to the strict MNN criteria and the inaccuracy of person detection.
We consider each segment different ID since there is no intra-camera tracklet label available.
We call our image sequence \emph{tracklet segments} to differentiate from the tracklets commonly used in the literature.

Together with a principal component reduction step using the camera labels, our method using the 480-hour dataset significantly outperforms existing methods that do not use any training data of the testing domain.
Our ablation study shows that tracklet accuracy has significant impact on the performance and the data size also has a positive effect on the ReID accuracy.
Therefore, there is great room to improve further with more advanced person detection and tracking algorithms, as well as with even larger and more diversified datasets.

The main contributions of this paper include:
\begin{itemize}[noitemsep,topsep=0pt,parsep=0pt,partopsep=0pt]
    \item an unsupervised ReID algorithm that applies contrastive learning in a novel fashion, first at the image level and then at the tracklet segment level, to achieve state-of-the-art performance.
    \item the first experimental report of unsupervised ReID with hundreds of hours of unlabeled pedestrian video, which demonstrates logarithmic performance increase as data size increases.
    \item the first unsupervised ReID algorithm to achieve top-tier accuracy without using any training data of the testing domain.  We train the model only once and apply it to all benchmarks.
\end{itemize}

\section{Related Work}\label{sec:related}
Person ReID has a long research history, from classical image processing methods to more recent deep learning based solutions~\cite{zheng2016person,karanam2019systematic}.
Here we focus on unsupervised methods, which can be largely divided into unsupervised ReID that uses only unlabeled training data of the testing domain, and unsupervised domain adaptation~(UDA) ReID that also includes a labeled dataset from a source domain in training.

\noindent \textbf{Unsupervised ReID}~~
methods typically start by assigning each person image or tracklet to its own class and gradually merge them by distance in the feature space, including nearest neighbor~\cite{lin2019bottom_BUC,lin2020unsupervised,wu2020tracklet_TSSL}, mutual nearest neighbor~\cite{li2020unsupervised_UTAL,wu2019unsupervised_UGA,zhu2019intra_MTML,yang2019patch_PAUL,wang2020unsupervised_MMCL}, average distance~\cite{ding2019dispersion_DBC,zeng2020hierarchical_HCT}, k-means~\cite{yu2020unsupervised_DECAMEL}, and DBScan~\cite{ge2020self}.
Once merged, we can assign pseudo or soft labels to these images to train the network using soft cross entropy~\cite{lin2019bottom_BUC,ding2019dispersion_DBC,lin2020unsupervised}, triplet~\cite{yang2019patch_PAUL,zeng2020hierarchical_HCT}, or contrastive losses~\cite{ge2020self}.

Because existing datasets are small and relatively homogeneous, these algorithms are carefully tuned toward these datasets for the best performance.
For example, many methods employ hyper parameters to control the size of the clusters, which is very close between Market and Duke.
When the size does not match the statistics of the datasets, the performance can drop by almost half or the training may not converge~\cite{zeng2020hierarchical_HCT,wang2020unsupervised_MMCL}.
With tracklet-based methods, the top performing algorithms assume at most one tracklet per ID per camera view, which is always true for image ReID datasets because all images of the same ID under a camera is considered one tracklet~\cite{zhu2019intra_MTML,wu2019unsupervised_UGA}.


\noindent \textbf{Unsupervised domain adaptation ReID}~~
employs similar strategy of clustering and pseudo label assignments~\cite{fu2019self,zhong2019invariance,yu2019unsupervised,ge2020mutual,ge2020self}.
Assisted by labeled training data in a source domain, there are more versatile and accurate label assignment techniques, including body parts models~\cite{fu2019self}, person attributes~\cite{wang2018transferable}, reference persons in the source domain~\cite{yu2019unsupervised}, invariance losses~\cite{zhong2019invariance} and similarity consistency losses~\cite{wu2019unsupervised} in the target domain.

While our problem setting also includes a source domain for training and a target domain for testing, we differ from UDA fundamentally since our source domain is unlabeled and we do not involve target domain data in training.

CycAs is the only ReID method we know that trained models by an unlabeled source only, a 6-hour video dataset in high pedestrian density areas~\cite{wang2020cycas}.
The method requires significant cross-camera view overlap with accurate timestamps to correlate frames to achieve good performance.
Despite the strong requirements, the testing results using 6-hour video falls significantly behind our method.

Finally, we adopt the contrastive learning framework from image classification that showed results comparable to supervised learning using \emph{contrastive instance discrimination}~\cite{he2020momentum_moco,chen2020big_simclrv2}.
Here we apply the contrastive loss in a novel fashion for the ReID problem, first at the instance level similar to image classification and then at the tracklet segment level for better ReID performance.

\section{Our Method}
This section discusses our unsupervised ReID algorithm in detail.
\subsection{Overview}
\begin{figure*}
    \centering
    \includegraphics[width=\textwidth]{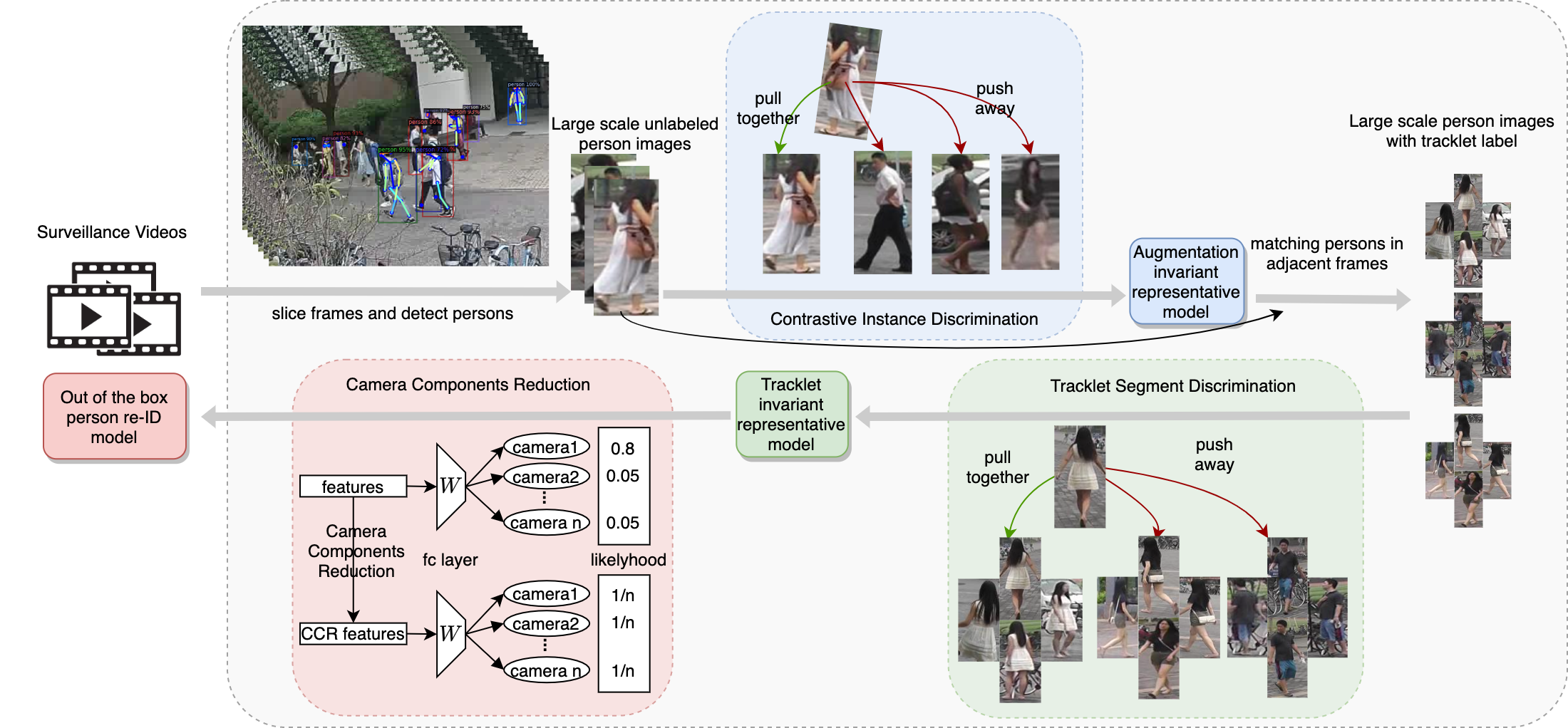}
    \caption{The overview of our method. 
    First we extract person bounding boxes using an object detection tool and apply contrastive instance discrimination~(CID) on these images to learn a base ReID model.
    Next we apply the base model to all the bbox images to obtain their feature vectors, which are used to aggregate mutual nearest neighbors of consecutive frames into tracklet segments.
    We apply contrastive learning again on these tracklet segments to obtain the second iteration model, called tracklet segment discrimination~(TSD).
    Since there is no inter-camera tracklet association, our model inevitably learns features that discriminate camera views.
    Therefore, our last step aims at eliminating these features through principal component analysis using camera labels freely available in our dataset, called camera components reduction~(CCR) step.
    There is no additional training involved in this step.}
    \label{fig:overview}
\end{figure*}

As Fig.~\ref{fig:overview} illustrates, we first slice the video into frames and detect pedestrians in each frame using object detection algorithms.
Next we apply contrastive learning algorithms on these extracted images to learn a base model, called contrastive instance discrimination~(CID).
We use this model to find tracklet segments in the next step, which associates images of consecutive frames if their feature vectors calculated using the base model are mutual nearest neighbors among all the images between the two frames.
Applying contrastive learning on these tracklet segments gives much better performance on ReID benchmarks than the base model.
We further improve its performance by removing components most relevant to camera views using principal component analysis and camera labels.

Before diving into details of our algorithm, we formulate our problem as follows.
Given unlabeled surveillance videos $\{ V_1, V_2,... \}$, we aim to train a representation model $f(x;\Phi)$ with weight $\Phi$ that for any query person image $x_i$, it aims to satisfy:
\begin{align*}
    \max\limits_{x'}(\{\text{dis}(f(x_i), f(x_i')\}) < \min\limits_{j}(\{\text{dis}(f(x_i), f(x_{j\ne i})\})
\end{align*}
where $x_i'$ is an image of the same ID of $x_i$ and $x_{j\ne i}$ is an image of a different ID. 
$\text{dis}(*,*)$ is the distance between the two feature vectors, which we use Euclidean distance.


\subsection{Contrastive Instance Discrimination}
\label{CID}
The CID step first processes the video to get person images. 
We pick the frames from the videos at a fixed interval, typically 2 fps, and then apply an off-the-shelf object detection tool which is based on Faster-RCNN to detect all pedestrian bounding boxes in each frame. 
This gives us a large set of unlabeled person images.

Unsupervised contrastive learning performs on par with supervised learning in recent image classification literature~\cite{he2020momentum_moco,chen2020big_simclrv2}, which we apply directly in our CID step.
More specifically, contrastive learning is a paradigm that learns distinctive representations from data organized into positive and negative sets.
With contrastive instance discrimination, the positive set includes an image and its augmented versions, while the negative set is all other images.
We use the widely adopted InfoNCE loss to train our model, defined as:
\begin{equation}\label{eq:infoNCE}
\mathcal{L}_{q, k^{+},\{k^{-}\}} = -\log \frac{\exp ( s_{q, k^{+}}  / \tau)}{\exp (s_{q, k^{+}}/ \tau)+\sum_{k^{-}} \exp  (s_{q , k^{-}}  / \tau) }
\end{equation}
where $s_{q,k}$ is the cosine similarity between $\boldsymbol{q}$ and $\boldsymbol{k}$, calculated by $
s_{q,k} =  {\boldsymbol{q} \cdot \boldsymbol{k}}/({|\boldsymbol{q}| \cdot |\boldsymbol{k}|})$. $\boldsymbol{q}$ is a query representation of an instance. $\boldsymbol{k^+}$ and $\boldsymbol{k^-}$ are the representations of positive and negative instances, respectively. 
$\tau$ is a temperature hyperparameter.
Therefore, the InfoNCE loss maximizes the similarity between the query representations and its positive representations while minimizing the similarity between the query representation and all the negative representations. 

In practice, we cannot include all negative distances in one batch due to memory limitation and the slow convergence.
But it is crucial to include a large enough negative dataset for better performance.
To handle this problem, SimCLR~\cite{chen2020big_simclrv2} chooses to enlarge the mini-batch size of training using large-scale parallelism, while MoCo~\cite{he2020momentum_moco} is based on more standard hardware utilizing a memory bank with momentum update to achieve similar results. 
We adopt the MoCo learning framework for our method.

It is straightforward to apply contrastive instance discrimination with the InfoNCE loss to the person images extracted, but the model performs poorly on ReID benchmarks in our experiments as well as reported in the literature~\cite{ge2020self}.
This suggests that the model is not learning ID discriminative features.
Nevertheless it performs much better than ImageNet pretrained models.
Therefore we use this model to find tracklet segments in the next step, which effectively enlarges the positive set.

\subsection{Tracklet Segment Discrimination}\label{sec:method-tracklet}
\label{TSD}
Unsupervised person ReID methods usually mine the spatial-temporal information of the video data, that is, person appeared in some frames would also appear in the near frames and near positions. 
We call it a tracklet that tracking some person through several frames. 
In this section we generate the tracklet label by feature extracted from CID trained model and then train the model using tracklet segment discrimination~(TSD) to learn a tracklet invariant representation model.


We do tracklet association using feature extracted by the CID trained model. 
For every video with frames $\{F_1, F_2,...\}$ and for any frame $F_i$ there are $n_i$ person images $\{ x^i_1, x^i_2,...x^i_{n_i}\}$. 
We calculate the affinity matrix between $F_i$ and $F_{i+1}$ by multiplying two feature matrix of the frame, to get a $n_i\times n_{i+1}$ matrix $A_{n_i\times n_{i+1}}$. 
Every element $a_{m,k}$ of $A_{n_i\times n_{i+1}}$ is $s_{f(x^i_m), f(x^{i+1}_k)}$ as calculated in Eq.~\ref{eq:infoNCE}. 

We adopt a \textit{mutual nearest neighbor} strategy to generate tracklets. 
In every Affinity matrix, if some element $a_{m,k}$ is the maximum in its row and also in its column, it means $x^i_m$ and $x^{i+1}_k $ are mutual nearest neighbors, then we call it a match. 
Every match will continue a tracklet to the new frame, and every unmatched image in $F_i$ is an end of a tracklet while every unmatched image in $F_{i+1}$ is a start of a tracklet. 
After the tracklet generation for all the videos, we get labels for TSD.


In the TSD step, we adopt the same contrastive learning procedure as Sec.~\ref{CID}. 
In CID, we aim to learn augmentation invariant representation model, so we treat the same instance with different augmentations as positive samples and other instances as negative samples. 
Now in TSD, we aim to learn tracklet invariant representation model, so we treat images from the same tracklet as positive samples and from other tracklets as negative samples.

By contrastive learning to discriminate tracklets, we train the model to be tracklet invariant enough for person ReID tasks.
There exist some tracklet-based methods for unsupervised person ReID, such as UTAL\cite{li2020unsupervised_UTAL} and TAUDL\cite{li2018unsupervised_TAUDL}, but they are not applicable in our large scale situation. 
They use set full-connected~(FC) layers for all the tracklets and use a multi-task strategy, which is to classify tracklets within each camera. 
It is not applicable because the FC layer would use up the graphic memory. 



\label{UTA}

\subsection{Camera Components Reduction}
For ReID task, we need a model that maps person images with any light conditions, camera intrinsics and other irrelevant variables from person ID to features that are close to each other in the embedding space. 
To achieve this, supervised methods rely on expensive human annotation that match persons from different cameras, while other unsupervised methods, e.g. UTAL, UGA, seek to design cross-camera matching methods to auto-generate cross-camera pseudo label. 
Inspired by principal component analysis, we propose a Camera Components Reduction~(CCR) method to eliminate the components that are irrelevant with person ID discrimination.



Principal components analysis is a common technique for dimension reduction. 
Given a centered matrix $W_{m\times n}$ that contains $m$ features of dimension $n$ and meets $m \leq n$, PCA seeks to do singular value decomposition that
\begin{equation}
    W_{m\times n} = U \Sigma V^{{\mathrm{T}}}
\end{equation} 
where $V^{\mathrm{T}}$ is a group of orthogonal basis that maps vectors from the $n$-dimensional space to new $m$-dimensional space;
$\Sigma$ is a diagonal matrix containing the eigen values of each orthogonal basis.
$\Sigma$ is sorted in decreasing order, meaning that the significance of new dimensions decreases by order. 
Conventional PCA uses the first $k$ orthogonal basis as the principal components and reduces the dimension by projecting original data to new $k$-dimension space using first $k$ rows of the $V^{{\mathrm{T}}}$.

Now, supposing $W_{m\times n}$ is the weight of a linear classifier for $m$-camera classification with feature embedding size $n$. 
By PCA we find the most significant $k$ dimensions for camera classification, which contains the encoded information of camera view points, camera intrinsics and other ID-irrelevant variables. 
The principal components for camera-discrimination are the irrelevant components for ID-discrimination.

We re-porject the $k$-dimensional vector back to the original $n$-dimensional space and eliminate it from the original embedding feature. 
Given any embedding feature $F$ with shape $n\times 1$, the irrelevant components eliminated feature $F^{\text{CCR}}$ is calculated by:
\begin{equation}\label{eq:CCR}
    F^{\text{CCR}} = F - V V^{\mathrm{T}} F   = (I - V V^{\mathrm{T}})F
\end{equation}
where $I$ is an identity matrix. After CCR, we make our tracklet-invariant representation feature generated in Sec.~\ref{TSD} to be camera-invariant, further improving the performance of our method.

Now we discuss why CCR makes the features camera-invariant.
When classifying the cameras, we multiply the feature $F$ with weight matrix $W$ to get an $m$-dimensional logit vector and then use the Softmax function to get the probability distribution:
\begin{align}
    \text{logit} &= WF \\ 
   P(y=j|\text{logit} )&={\frac {e^{\text{logit}_{j}}}{\sum _{i=1}^{m}e^{\text{logit}_{i}}}}
\end{align}

After CCR, if we calculate the probability:
\begin{align}\begin{aligned}
    \text{logit}^{\text{CCR}} &= W\cdot F^{\text{CCR}} = U\Sigma V^{\mathrm{T}} (I - V V^{\mathrm{T}})F \\ 
    &= (U\Sigma V^{\mathrm{T}} - U\Sigma V^{\mathrm{T}}V V^{\mathrm{T}} )F = \vec{0}
    \end{aligned}\\
    \begin{aligned}
   P(y=j|\vec{ 0} )&={\frac {e^0}{\sum _{i=1}^{m}e^0}} = \frac{1}{m}
    \end{aligned}
\end{align}
where $ V^{\mathrm{T}}\cdot V = I$ because $ V^{\mathrm{T}}$ is a group of orthogonal basis.
The classifier will classify any feature to have equal probabilities to all the cameras, meaning that the components for camera classification are eliminated.

\section{Experiments}
We discuss our detailed experimental settings and results in this section.

\subsection{Datasets}
We collect surveillance video from an enclosed campus as our training data. 
We pick 24 cameras from high pedestrian density areas such as dining halls, classroom buildings, and intersections.
We choose 4 hours of peak traffic time each day for five days, a total of 480 hour raw footage.
The preprocessing of our data is done by sampling the video at two frames per second using FFmpeg and then using Detectron2~\cite{wu2019detectron2} to extract person bounding boxes in every frame by Faster-RCNN~\cite{ren2015faster}. 
After filtering bounding boxes by resolution and confidence thresholds, we get about 15.4M bounding boxes. 

\begin{figure*}
    \centering
    \includegraphics[width=\textwidth]{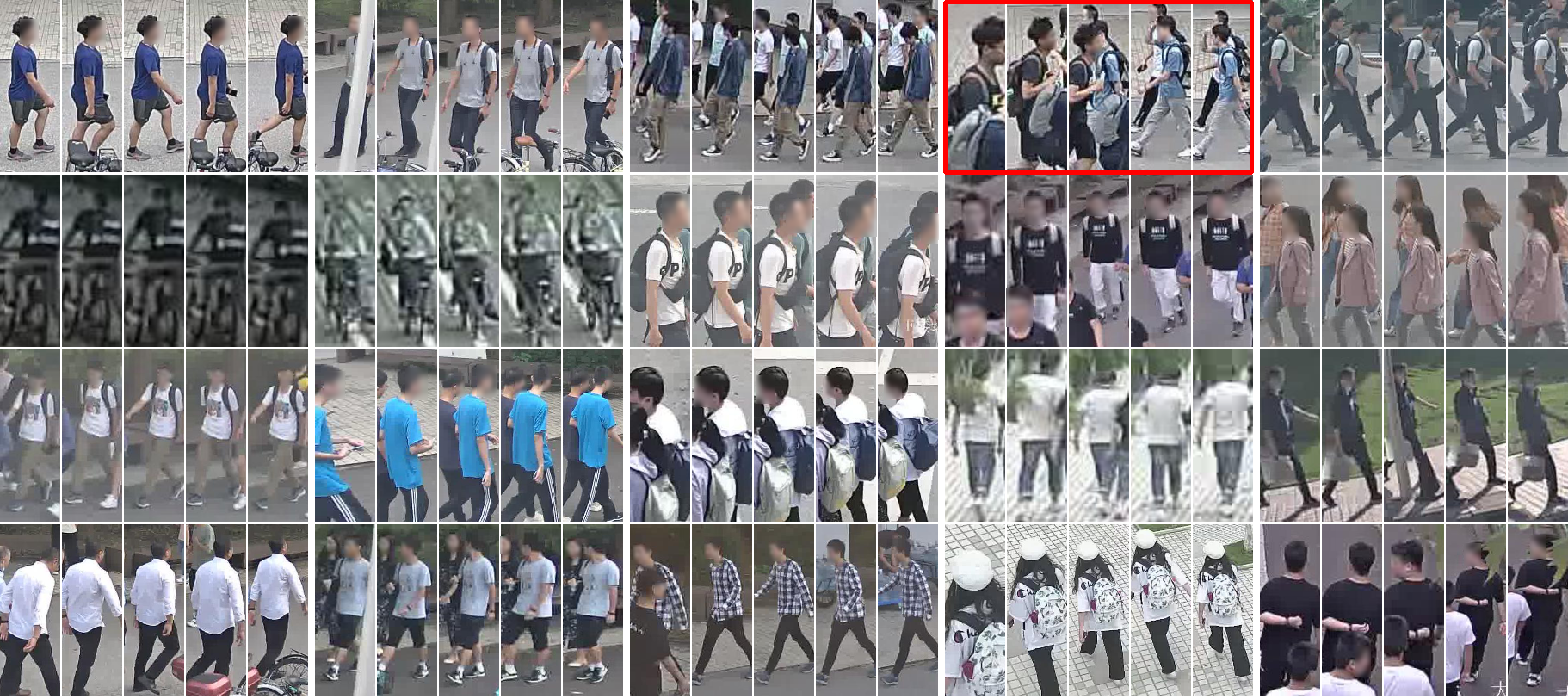}
    \caption{Samples tracklets in CampusT dataset. The tracklet in the red box is a wrong tracklet, resulting from the tracking errors when two man cross each other in the camera view.}
    \label{fig:datasample}
\end{figure*}

After training step CID, we extract the features of all the person images and match them by \textit{mutual nearest neighbor} between every two adjacent frames and get about 3M tracklets in total. 
Some samples are shown in Figure~\ref{fig:datasample}.
There are many errors in these auto-generated tracklet labels, which may cause different persons to appear in one tracklet and cause one tracklet to break down into multiple parts. 
The tracklet in the red box of Figure~\ref{fig:datasample} is a good example, in which we track the wrong person when two persons cross each other in the camera view. 
Since we focus on contrastive learning in this paper, we do not further optimize our tracklet extraction algorithm that can lead to additional performance gain.

\begin{figure}
    \centering
    \includegraphics[width=.49\textwidth]{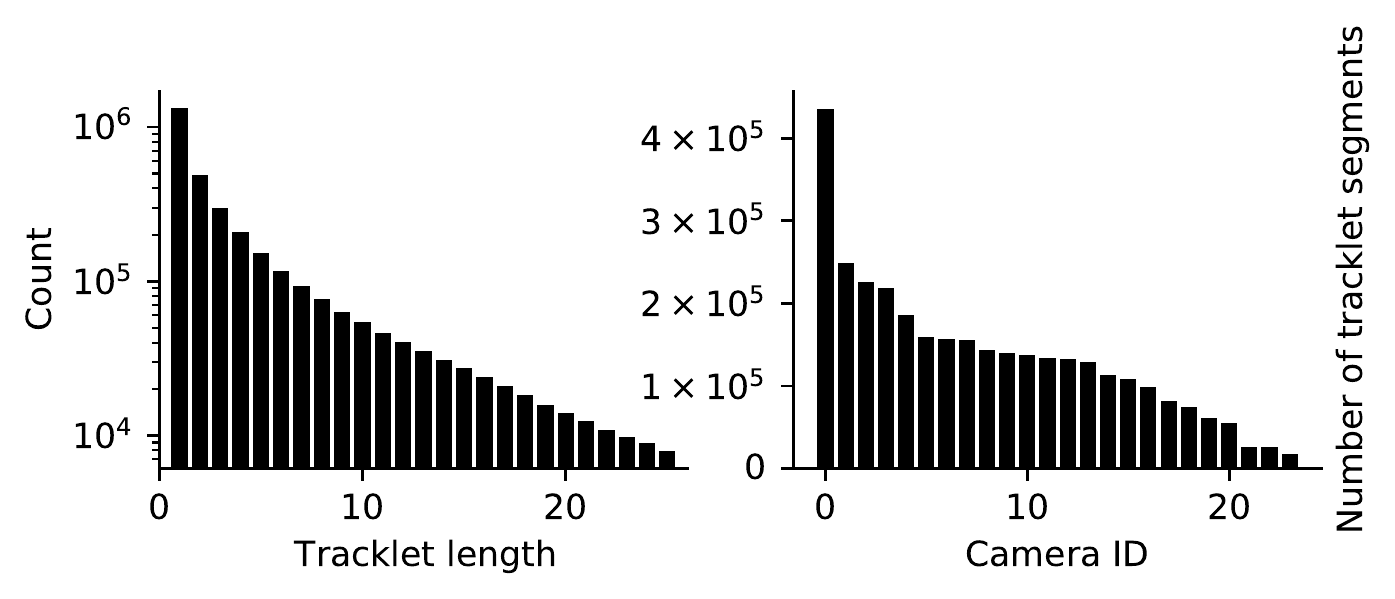}
    \caption{ \textbf{Left}: Histogram of tracklet counts by their lengths. 
    \textbf{Right}: Histogram of tracklet counts by different cameras. 
    }
    \label{fig:tracklet_length_count}
\end{figure}

The statistics of our tracklet lengths and their distribution across cameras is shown in Figure~\ref{fig:tracklet_length_count}, which follow typical long tail distribution.
Single-image tracklets make up 40.52\% of all the tracklets, and 71.37\% of the tracklets are shorter than 5, i.e., 2 seconds.
Removing the tracklets shorter than 2 seconds, we get 946K tracklets in total. The tracklets are distributed unevenly among all the cameras.

\subsection{Implementation Details}
In the training CID\&TSD steps, we use the MoCo implementation of contrastive learning as our training framework with its default setting, i.e., memory bank size 65536, temperature 0.07, learning rate 0.03, moco-dim 128, moco-momentum 0.999, and batch size 512.
We choose ResNet-50~\cite{he2016resnet} as our backbone for fair comparison to other methods.
We train our model with SGD optimizer on 8 NVidia-Tesla-A100 GPUs for most of our experiments.
For data augmentation, we adopt random resized crop, random horizontal flip, and random color jitter at training time and only resize at testing time.
We train 10 epochs for CID and train 50 epochs with cosine-annealing lr-scheduler for TSD. 
Both training steps take about 12 hours each.

We report CMC ranks and mAP as other person re-ID methods. 
Due to the lack of human annotation on our dataset, we can not evaluate the performance on our dataset. 
We aim to train the model the general representation of person-ID, so we pick the two commonly used Market-1501~\cite{zheng2015scalable_market} and DukeMTMC-reID~\cite{gou2017dukemtmc4reid} person re-ID datasets to evaluate. 
Comparing to most other ReID methods that report results on these benchmarks, the most salient difference is that we do not use any of their training data.

\subsection{Ablation Studies}
We perform ablation studies to discuss effects of tracklet labels, different steps of our method, data amount, model capacity, and the generalization of our method to adapt to other domains.

\paragraph{Model Performance vs. Tracklet Label Quality.}
As training data of the tracklet invariant representation model, tracklet is critical for model to learn distinctive feature. 
Since our tracklet association method is naive, the tracklet labels would inevitably contain a lot of noise.  
To discuss the effect of the tracklet label, we design experiments to use different noisy levels of tracklet label.

\begin{figure}
    \centering
    \includegraphics[width=.45\textwidth]{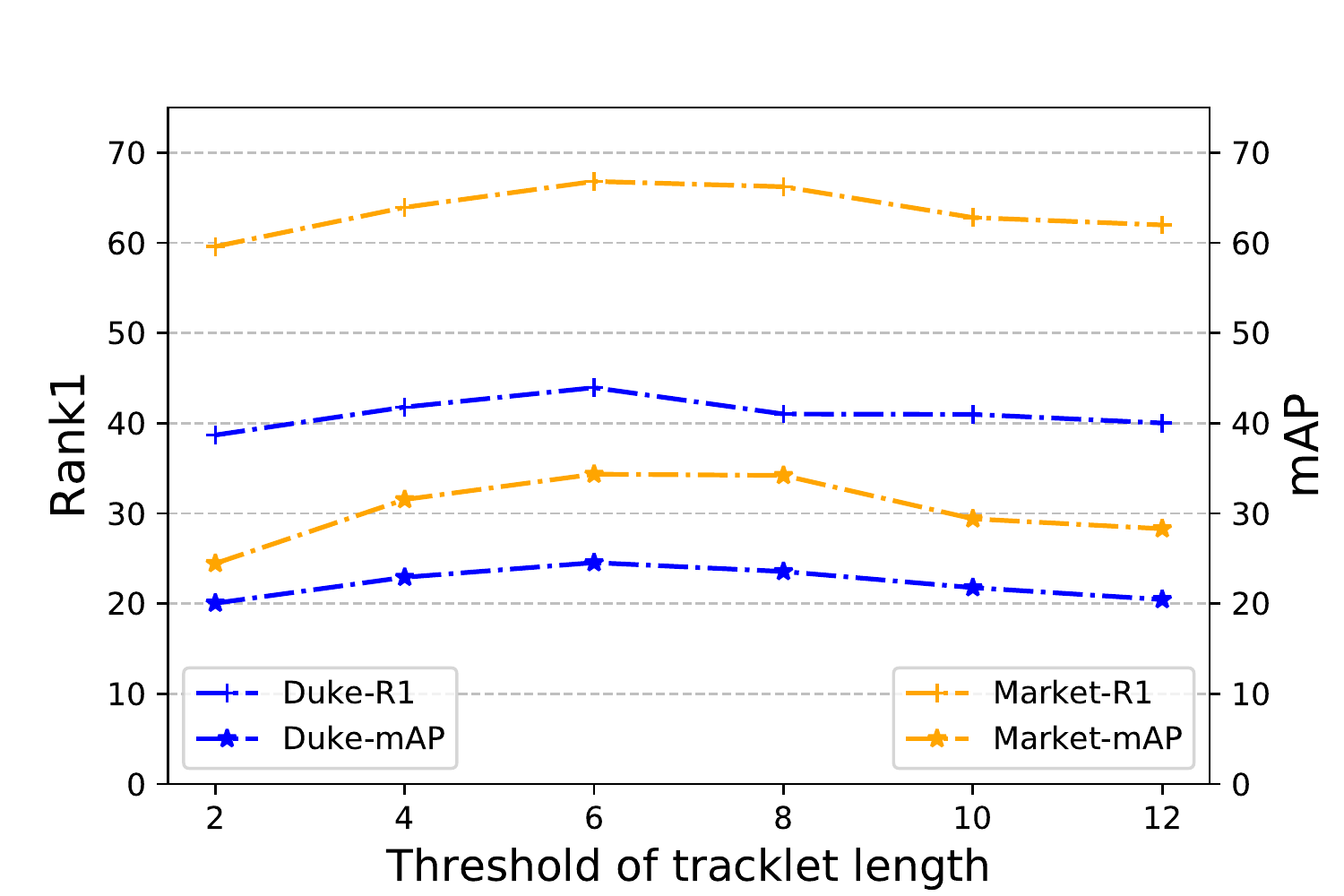}
    \caption{Performance with different tracklet length thresholds. We train the TSD model with the tracklets that are longer than the thresholds.}
    \label{fig:tracklet_len}
\end{figure}

Our one strategy is to use different thresholds to filter the short tracklets, and with the higher threshold, the tracklets will be more precise and their counts will be less. For a fast comparison, we use only 10\% of the total dataset. The result is shown in Fig~\ref{fig:tracklet_len}. As the thresholds grow, the performance increase and then drop gradually and the plateau is the best trade-off between the tracklet qualities and quantities.

We also set an experiment that use a more mature object tracking method to match the person images and get more accurate tracklet labels. Tab~\ref{tab:tracking_method} shows the results, and we observe that the performance increase with better tracking methods.


\begin{table}[]
\centering
\resizebox{.49\textwidth}{!}{
\begin{tabular}{c|cc|cc}
\hline
\multirow{2}{*}{Tracking Method} & \multicolumn{2}{c|}{Market} & \multicolumn{2}{c}{DukeMTMC-ReID} \\ \cline{2-5} 
                          & Rank-1       & mAP        & Rank-1              & mAP               \\ \hline
Our MNN & 66.8         & 34.4       & 43.7                & 24.5              \\
TRMOT\cite{wang2019towards}      
& 68.0         & 35.2       & 43.9                & 25.0             \\ \hline
\end{tabular}}
\vspace{1ex}
\caption{Performance using different tracking methods to generate tracklets, using 10\% of the data.}
\label{tab:tracking_method}
\end{table}


\paragraph{Model Performance vs. Training steps.}

\begin{table}[htb]
\centering
\resizebox{.49\textwidth}{!}{
\begin{tabular}{c|c|c|cc|cc}
\hline
\multirow{2}{*}{CID} & \multirow{2}{*}{TSD} & \multirow{2}{*}{CCR} & \multicolumn{2}{c|}{Market} & \multicolumn{2}{c}{DukeMTMC-ReID} \\ \cline{4-7} 
                     &                      &                      & Rank-1        & mAP         & Rank-1            & mAP            \\ \hline
\checkmark           &                      &                      & 25.1          & 6.3         & 19.7              & 6.3            \\
                     & \checkmark           &                      & 57.6          & 28.1        & 35.2              & 19.7           \\
\checkmark           & \checkmark           &                      & 68.9          & 36.4        & 49.4              & 28.9           \\
\checkmark           & \checkmark           & \checkmark           & 72.7          & 40.0        & 51.7              & 31.2           \\ \hline
\end{tabular}}
\vspace{1ex}
\caption{Effectiveness of different steps of our method.}
\label{tab:steps}
\end{table}

Table~\ref{tab:steps} shows the effectiveness of different steps of our method.

The most notable contribution of CID is to train an augmentation invariant representation model to match person images between adjacent frames and get tracklet labels. 
But there are more it can do other than auto-labeling. 
Through CID, the model learns lots of hidden representation of the instances that can not be learned through other steps, which contributes to the performance.
Comparing experiment 2 to 3 in Tab.~\ref{tab:steps}, we find that a model trained with TSD only would suffer a great performance decrease, e.g. rank-1 drops from 68.9 to 57.6 in Market. 
It proves that even if we could annotate the data using a pre-trained ReID weight, the CID is still necessary for the model to learning extra representations to distinguish the person IDs.

TSD contributes most in our pipeline, through which the performance gets the highest improvement. 
By TSD, we train our model to contrastively discriminate among tracklets and pull together features that belong to the same tracklet. 
It is worth noticing that we did not apply the more sophisticated multi-object tracking method, so the tracklet quality is not so good. 
Although the quantity of data is huge, it is still challenging to learn good ID-representations through such low level auto-generated labels. 
Comparing experiment 1 to 3 or 1 to 2 in Tab.~\ref{tab:steps}, the performances increase greatly through TSD on both two datasets.

We further improve the performances by eliminating the known irrelevant camera information through CCR procedure.  
Comparing experiment 3 to 4 in Tab.~\ref{tab:steps}, the CCR consistently improves the performances on both datasets. 
On Market and DukeMTMC-ReID, the performances increase 3.8 and 2.3 on Rank-1 respectively, and this is consistent to the absolute performances on these datasets. 
We believe the domain gap from our data to DukeMTMC-ReID is larger than from Market, and that is why the performance gains of CCR are different in the two datasets.

\begin{figure}
    \centering
    \includegraphics[width=.49\textwidth]{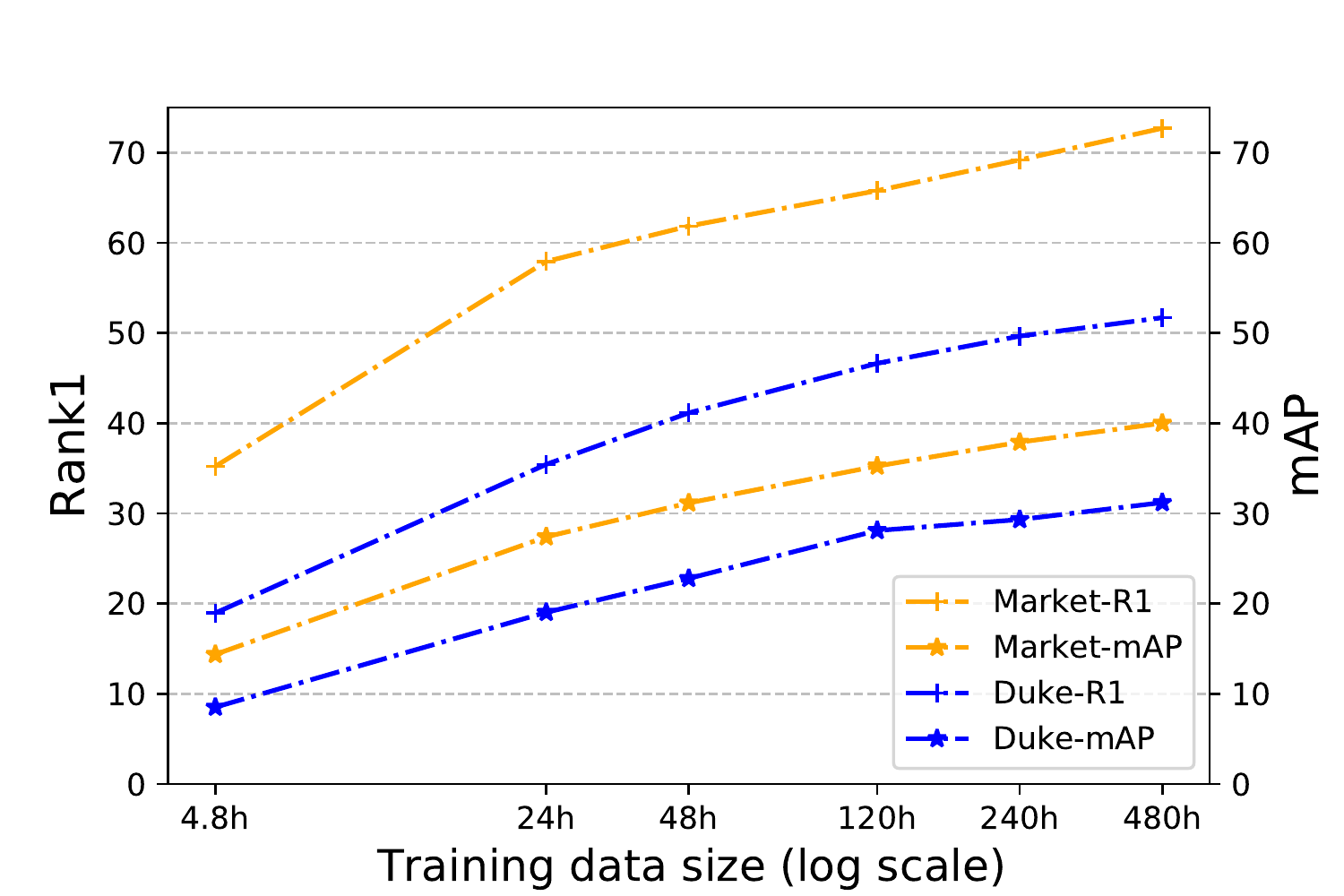}
    \caption{Performance with different data sizes. We train our model with 1\%, 5\%, 10\%, 25\%, 50\%, 100\% of our data. As the data size grows, the performance improves continually.}
    \label{fig:ablation_datasize}
\end{figure}

\paragraph{Model Performance vs. Data Size}
Our method is designed to train models to learn person ID representations by easy-to-get unsupervised videos. 
So the scalability to data sizes is critical for our method. 
To discuss the scalability, we design experiments to use parts of our 480 hours person image dataset. 
We split our dataset not randomly but by cutting the video length, to simulate the real situations of enlarging the data scale. 
We set experiments to use 1\%, 5\%, 10\%, 25\%, 50\% and 100\% of our total data and evalute the performance on Market and DukeMTMC-ReID datasets. 
The performance is shown in Figure~\ref{fig:ablation_datasize}, it shows that as the data scale grows, the performances improve continually without saturation. 
So our method can achieve better performances with even larger data. 
We believe with data from more complicated scenes and of larger size, our method can train a model that can extract the most general person ID representation and perform well on every test dataset.

\begin{figure}
    \centering
    \includegraphics[width=.5\textwidth]{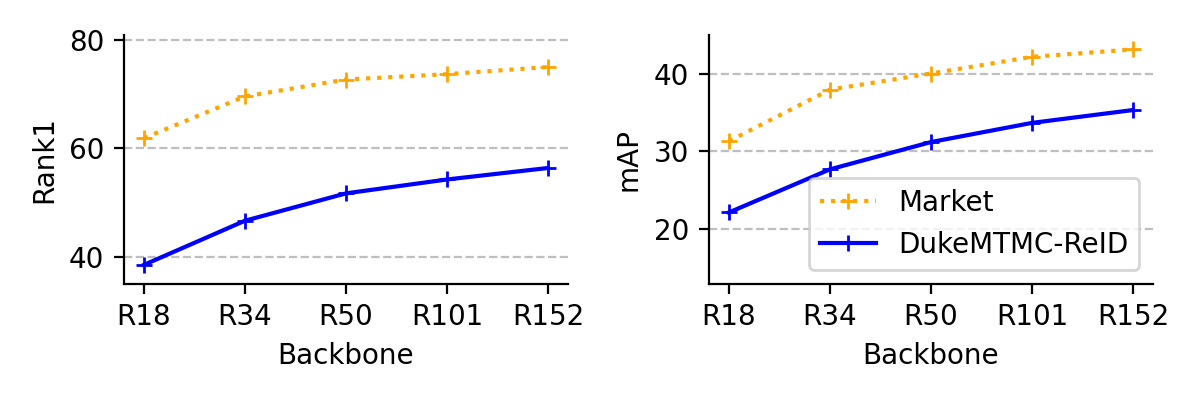}
    \caption{Performance with increasing model sizes, i.e., ResNet-18, -34, -50, -101 and -152. 
    }
    \label{fig:ablation_modelsize}
\end{figure}



\begin{table*}[!ht]
\centering
\resizebox{\textwidth}{!}{
\begin{tabular}{c|c|cccc|cccc}
\hline
\multirow{2}{*}{Method}                                          & \multirow{2}{*}{Training Data}                                                     & \multicolumn{4}{c|}{Market-1501} & \multicolumn{4}{c}{DukeMTMC-reID} \\ \cline{3-10} 
                                                                 &                                                                                    & Rank-1 & Rank-5 & Rank-10 & mAP  & Rank-1  & Rank-5  & Rank-10 & mAP  \\ \hline \hline
Supervised~\cite{Luo2019bag}                                                      & labeled data from testing                              & 95.4& -             & -             & {94.2} & {90.3} & -             & -             & {89.1} \\ \hline 
SPGAN~\cite{deng2018image}                                       & \multirow{6}{*}{\shortstack{labeled data from \\ a source domain, \\ unlabeled data from \\ the testing domain \\ (i.e., unsupervised \\ domain adaptation)}} & 57.0   & 73.9   & 80.3    & 27.1 & 41.1    & 56.6    & 63.0    & 22.3 \\
HHL~\cite{zhong2018generalizing}                                 &                                                                                    & 62.2   & 78.8   & 84.0    & 31.4 & 46.9    & 61.0    & 66.7    & 27.2 \\
MAR~\cite{yu2019unsupervised}                                    &                                                                                    & 67.7   & 81.9   & 87.3    & 40.0 & 67.1    & 79.8    & 84.2    & 48.0 \\
ENC~\cite{zhong2019invariance}                                   &                                                                                    & 75.1   & 87.6   & 91.6    & 43.0 & 63.3    & 75.8    & 80.4    & 40.4 \\
SSG~\cite{fu2019self}                                            &                                                                                    & 80.0   & 90.0   & 92.4    & 58.3 & 73.0    & 80.6    & 83.2    & 53.4 \\
MMT~\cite{ge2020mutual}                                            &                                                                                    & \textbf{87.7}   & \textbf{94.9}   & \textbf{96.9}    & \textbf{71.2} & \textbf{78.0}    & \textbf{88.8}    & \textbf{92.5}    & \textbf{65.1} \\ \hline
Wu~\etal~\cite{wu2019progressive}                                & \multirow{8}{*}{\shortstack{unlabeled data from \\ the testing domain }}                    & 55.8   & 72.3   & 78.4    & 26.2 & 48.8    & 63.4    & 68.4    & 28.5 \\
DECAMEL~\cite{yu2020unsupervised_DECAMEL}                        &                                                                                    & 60.2   & 76.0   & -       & 31.4 & -       & -       & -       & -    \\
BUC~\cite{lin2019bottom_BUC}                                     &                                                                                    & 61.9   & 73.5   & 78.2    & 29.6 & 40.4    & 52.5    & 58.2    & 22.1 \\
PAUL~\cite{yang2019patch_PAUL}                                   &                                                                                    & 68.5   & 82.4   & 87.4    & 40.1 & 72.0    & 82.7    & 86.0    & 53.2 \\
UTAL~\cite{li2020unsupervised_UTAL}                              &                                                                                    & 69.2   & -      & -       & 46.2 & 62.3    & -       & -       & 44.6 \\
HCT~\cite{zeng2020hierarchical_HCT}                              &                                                                                    & 80.0   & 91.6   & 65.2    & 56.4 & 69.6    & 83.4    & 87.4    & 50.7 \\
MetaCam+DSCE~\cite{yang2021joint}                                &                                                                                    & 83.9   & 92.3     & -       & 61.7 & 73.8    & 84.2      & -       & 53.8 \\ 
MTML~\cite{zhu2019intra_MTML}                                    &                                                                                    & 85.3   & -      & 96.2    & 65.2 & 71.1    & -       & 86.9    & 50.7 \\
UGA~\cite{wu2019unsupervised_UGA}                                &                                                                                    & 87.2   & -      & -       & 70.3 & 75.0    & -       & -       & 53.3 \\ \hline \hline
\multirow{2}{*}{\shortstack{Direct transfer\\~\cite{fu2019self,wang2020cycas}}} & \multirow{2}{*}{\shortstack{labeled data from \\ a source domain }}           & 54.6   & 71.1   & 77.1    & 26.6 & 30.5    & 45.0    & 51.8    & 16.1 \\
                                &         & 43.2   & -      & -       & 19.8 & 26.3    & -       & -       & 13.1 \\ \hline
BUC~\cite{lin2019bottom_BUC,wang2020cycas}                       & \multirow{3}{*}{6-hour unlabeled video}                                            & 29.8   & -      & -       & 14.2 & 21.5    & -       & -       & 11.2 \\
UGA~\cite{wu2019unsupervised_UGA,wang2020cycas}                  &                                                                                    & 37.2   & -      & -       & 17.8 & 25.6    & -       & -       & 15.4 \\
CycAs~\cite{wang2020cycas}                                       &                                                                                    & 50.8   & -      & -       & 23.3 & 34.6    & -       & -       & 19.2 \\ \hline
Ours                                                             & 480-hour unlabeled videos                                                           & \textbf{72.7} & \textbf{83.7} & \textbf{88.1} & \textbf{36.2} & \textbf{51.7} & \textbf{65.4} & \textbf{71.5} & \textbf{31.2} \\ \hline \hline
\end{tabular}

}
\vspace{1ex}
\caption{Comparison with other unsupervised methods.
The first section separated by double lines include the unsupervised methods that use the training data in the testing domain, include supervised, unsupervised domain adaptation, and unsupervised ReID methods. The second section includes methods that do not use any testing domain data.  Our method performs the best in this category. 
}\label{tab:sota}
\end{table*}

\paragraph{Model Performance vs. Model Size}
We use ResNet-50 as our backbone for a fair comparison to other methods. 
But in real cases, we may use better base extractors for better performances, so it is also important for our method to have scalability to the base model size. 
We set experiments of using ResNet series backbones, including ResNet-18, ResNet-34, ResNet-50, ResNet-101 and ResNet-152. 
As shown in Figure~\ref{fig:ablation_modelsize}, our model's performance increases by applying larger backbones, proving the scalability to model size of our method.

\subsection{Comparison with State-of-the-art Methods}

Table~\ref{tab:sota} shows the comparison with state-of-the-art unsupervised ReID methods.
Separated by double lines, the first section includes methods that use the training data provided by the benchmark datasets.
We further break these into three subsections, supervised baseline, unsupervised domain adaptation, and unsupervised methods.
As stated in Section~\ref{sec:related}, all the unsupervised methods rely on pseudo labels assigned by clustering.
Their results are approaching the limit of supervised training.

The second section presents methods that do not use the provided training data, not even in an unsupervised fashion.
We further break this down to direct transfer methods that train the model by a different labeled dataset and then test on the benchmark without any tuning, another study that utilized a private dataset of 6-hour hand-held video~\cite{wang2020cycas}, and our method trained by our 480-hour video.
Our method outperforms all the others by a significant margin in this category.

The performance of our method is worse with the DukeMTMC dataset than with Market-1501.
This is likely because our dataset and Market-1501 share similar characteristics such as student race distribution, clothes preferences, and campus environments.
We believe our performance on DukeMTMC can improve tremendously if there is a large set of video obtained in similar environments.

\section{Conclusion}
To the best of our knowledge, this is the first unsupervised ReID method that utilizes hundreds of hours of unlabeled pedestrian video.
It is also the first method that does not involve any testing domain data in training to achieve performance on par with state-of-the-art unsupervised methods.
Contrastive learning is the core ingredient of our method, which we apply in a novel way to both person images and tracklet segments.
These images and segments are very weak labels obtained by an off-the-shelf tool and a simple clustering rule, respectively.
The simplicity of our algorithm and the exclusion of testing domain data imply great application potential. 

{\small
\bibliographystyle{ieee_fullname}
\bibliography{egbib}
}

\end{document}